\documentclass[11pt]{article}

\usepackage[final]{acl}
\usepackage{booktabs}
\usepackage{booktabs} 
\usepackage{times} 
\usepackage{latexsym}
\usepackage{tabularx} 
\usepackage{array} 
\usepackage{enumitem} 
\usepackage[T1]{fontenc}

\usepackage[utf8]{inputenc}
\usepackage{makecell}
\usepackage{xcolor}
\definecolor{passgreen}{HTML}{2E7D32}
\definecolor{failred}{HTML}{C62828}
\definecolor{blockedgray}{HTML}{616161}
\definecolor{warnorange}{HTML}{EF6C00}
\usepackage{microtype}

\usepackage{inconsolata}

\usepackage{graphicx}
\usepackage{amssymb}
\usepackage{pifont}
\usepackage{xcolor}
\usepackage{pifont}

\definecolor{passgreen}{HTML}{2E7D32}
\definecolor{failred}{HTML}{C62828}
\definecolor{partialyellow}{HTML}{F9A825}

\newcommand{\cmark}{\textcolor{passgreen}{\ding{51}}}
\newcommand{\rmark}{\textcolor{failred}{\ding{55}}}
\usepackage{tikz}
\newcommand{\ymark}{
\tikz[baseline=-0.6ex]
\draw[partialyellow, line width=1.2pt] (0,0) circle (0.55ex);
}
\usepackage[most]{tcolorbox}
\usepackage{xcolor}
\usepackage{listings}
\usepackage{tikz}
\usetikzlibrary{shadows}
\usepackage{algorithm}
\usepackage{algpseudocode}
\definecolor{boxPurple}{HTML}{8491E8} 
\definecolor{codebg}{HTML}{FFFFFF}
\usepackage{roboto-mono} 
\usepackage[T1]{fontenc}
\newcommand{\codefontsize}{\fontsize{8pt}{10.5pt}\selectfont}
\lstdefinelanguage{json}{
    basicstyle=\ttfamily\codefontsize, 
    columns=fullflexible,
    backgroundcolor=\color{codebg},    
    showstringspaces=false,
    commentstyle=\color{white},
    keywordstyle=\color{blue},
    stringstyle=\color{teal},
    breaklines=true,
    breakatwhitespace=true,
    frame=none, 
    literate=
     *{0}{{{\color{purple}0}}}{1}
      {1}{{{\color{purple}1}}}{1}
      {2}{{{\color{purple}2}}}{1}
      {3}{{{\color{purple}3}}}{1}
      {4}{{{\color{purple}4}}}{1}
      {5}{{{\color{purple}5}}}{1}
      {6}{{{\color{purple}6}}}{1}
      {7}{{{\color{purple}7}}}{1}
      {8}{{{\color{purple}8}}}{1}
      {9}{{{\color{purple}9}}}{1}
      {:}{{{\color{black}:}}}{1}
      {,}{{{\color{black},}}}{1}
      {\{}{{{\color{black}\{}}}{1}     
      {\}}{{{\color{black}\}}}}{1}     
      {[}{{{\color{black}[}}}{1}
      {]}{{{\color{black}]}}}{1},
}

\newtcblisting{nicejson}[2][]{
    enhanced,
    colback=codebg,        
    colframe=boxPurple,    
    colbacktitle=boxPurple,
    coltitle=white,        
    fonttitle=\Large\rmfamily,
    title={#2},            
    sharp corners,
    drop shadow={opacity=1, shadow xshift=1.5mm, shadow yshift=-1.5mm, color=boxPurple},
    listing only,
    listing options={language=json}, 
    #1
}

\title{STAGE-Claw: Automated State-based Agent Benchmarking \\ for Realistic Scenarios}

\author{
    Sirui Liang\textsuperscript{1,3,4,6}\thanks{\ Co-first authors, they contributed equally to this work.}, 
    Bohan Yu\textsuperscript{1,2,4,6}\footnotemark[1], 
    Peiyu Wang\textsuperscript{1,3,4},
    Shiguang Guo\textsuperscript{6}, 
    Wenxing Hu\textsuperscript{6}, 
    Pengfei Cao\textsuperscript{1,3}, \\
    \textbf{Jian Zhao}\textsuperscript{4,5}, 
    \textbf{Cao Liu}\textsuperscript{6}, 
    \textbf{Ke Zeng}\textsuperscript{6},
    \textbf{Xunliang Cai}\textsuperscript{6}, 
    \textbf{Kang Liu}\textsuperscript{1,3}\thanks{\ \ Corresponding author.} \\
    \fontsize{10pt}{6pt}\selectfont
    $^1$The Key Laboratory of Cognition and Decision Intelligence for Complex Systems, Institute of Automation,\\
    \fontsize{10pt}{6pt}\selectfont$^2$School of Advanced Interdisciplinary Sciences, University of Chinese Academy of Sciences, \\
    \fontsize{10pt}{6pt}\selectfont $^3$Chinese Academy of Sciences, Beijing, China, University of Chinese Academy of Sciences, \\
    \fontsize{10pt}{6pt}\selectfont$^4$Zhongguancun Academy, $^5$Zhongguancun Institute of Artificial Intelligence,
    \fontsize{10pt}{6pt}\selectfont$^6$Meituan \\
    \texttt{\small \{liangsirui2024, yubohan2025\}@ia.ac.cn, kliu@nlpr.ia.ac.cn}\\
}

\begin{document}
\maketitle
\begin{abstract}
Large language models are increasingly used to power personal agents for everyday applications, but evaluating these agents remains a challenge. 
Existing benchmarks still rely on sandboxed artifacts, static task design, and coarse scoring, which hinder scalability and limit progress toward reliable personal-agent evaluation.
This paper introduces STAGE-Claw, an automated framework for building and evaluating realistic personal-agent scenarios in state-based personal-computing environments.
Given a task hint, STAGE-Claw automatically creates and validates a realistic benchmark task with its environment, task prompts, ground truth, and related verification programs. Agents are then evaluated in realistic operating environments, where performance is measured by the correctness of the final system state rather than only the textual response.
Using STAGE-Claw, this paper creates a benchmark with 40 challenging real scenario agent tasks, evaluates 11 frontier models, and analyzes their task scores, costs, tool-call reliability, and common failure patterns. Overall, STAGE-Claw offers a scalable, state-based way to evaluate agents in realistic user scenarios. Code is available \href{https://github.com/LiangThree/STAGE-Claw.git}{here}.
\end{abstract}

\begin{figure*}
  \includegraphics[width=\textwidth]{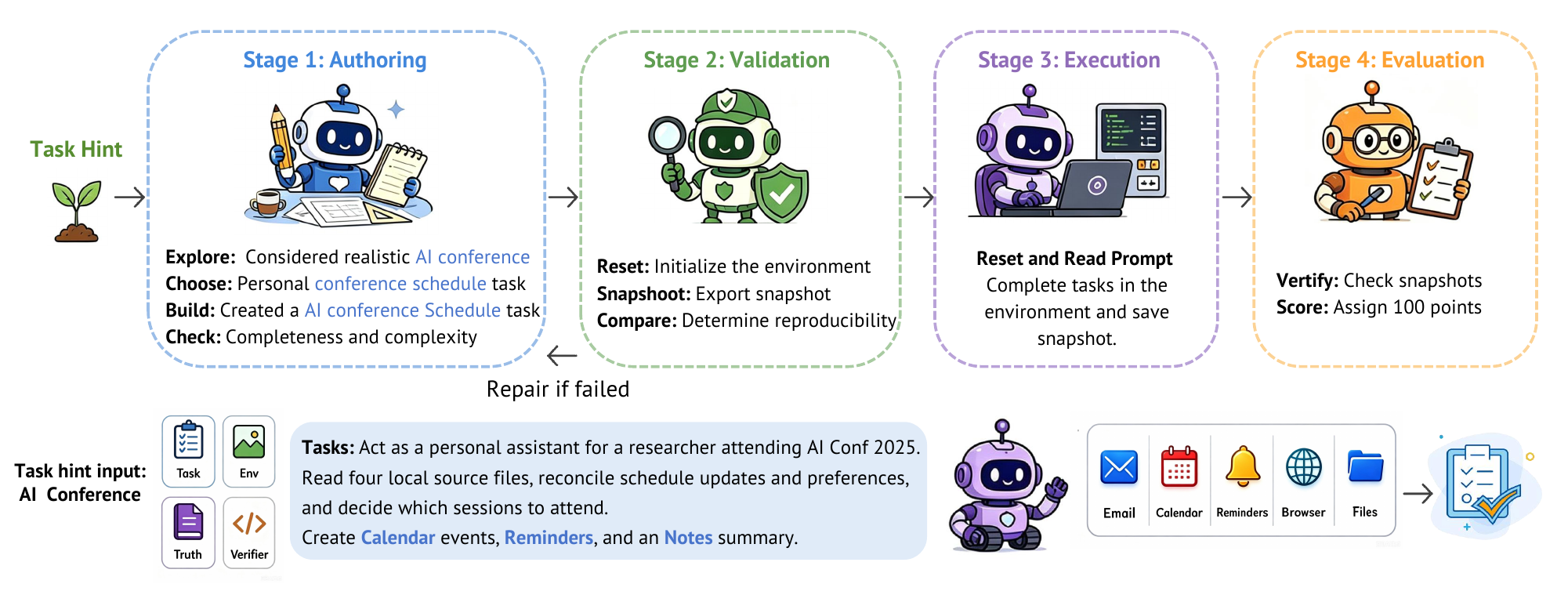}
  \vspace{-20pt}
  \caption{The workflow of STAGE-Claw. \textit{1) Benchmark authoring}: Explore a task hint and generate the task. \textit{2) Benchmark validation}: Check task correctness, difficulty, and reproducibility, revising if needed. \textit{3) Agent execution}: Target agent attempts the task in the environment. \textit{4) State-based evaluation}: Score results by verifying system state.}
  \label{fig:main}
\vspace{-10pt}
\end{figure*}

\section{Introduction}

Large language models (LLMs) are increasingly being adopted as the reasoning backbones of autonomous agents \citep{wang2024survey,xi2025rise,yao2022react}, as exemplified by systems such as Claude Code \cite{anthropic2025claude_code} and OpenClaw \cite{openclaw2026}, which augment LLMs with tool interfaces, execution environments, memory mechanisms, and control logic. 
This shift changes evaluation \cite{liu2024agentbench,ye2026claw,li2026clawsbench}, agents must not only answer textual prompts but also plan over multiple steps, coordinate heterogeneous tools, and interact with environments. 
For example, agent integrated with email, calendars, files, browsers, and other everyday applications \cite{openclaw2026}, benchmarks must measure reliable action in persistent, cross-tool user scenarios \cite{li2026clawsbench}.

Recent agent benchmarks have moved beyond text-only evaluation, covering tool use and multi-step reasoning, web instruction following, desktop interaction, and visually grounded web tasks~\citep{mialon2024gaia,deng2023mind2web,xie2024osworld,zhou2024webarena,koh2024visualwebarena}. 
However, they are still limited in three crucial aspects. 
\textbf{First, most existing evaluations replace real application state with sandboxed artifacts.} 
For example, PinchBench \cite{kilo2026pinchbench} evaluated calendar scheduling ability through generated \texttt{.ics} files and email-related tasks through synthetic inboxes stored as workspace text files. 
This file-based formulation simplifies task completion evaluation, but it could ignore some operations in real scenarios, such as software-permission and tool-access errors. 
Consequently, it mainly evaluates artifact generation rather than an agent's ability to interact with the applications. 
\textbf{Second, existing evaluation tasks are typically constructed manually and are therefore difficult to scale.}
GAIA \cite{mialon2024gaia} and Claw-Eval \cite{ye2026claw} rely on fixed question-answer instances or human-verified tasks and rubrics. However, personal agents must adapt to diverse user preferences, goals, workflows, and evolving contexts, requiring evaluation tasks that scale across personalized and dynamic scenarios. Such coverage is difficult and costly to achieve with manually curated static benchmarks. 
\textbf{Third, existing evaluations often lack process-aware diagnostics.} 
Only final results scoring or checkers \cite{ma2024agentboard, trivedi2024appworld} may fail to diagnose where errors occur within the completion workflow. 
For instance, an incorrect conference-tracking related calendar operation produced by an agent may result from errors in intermediate steps, such as time-zone conversion, conflict resolution, or reconciling inconsistent message sources, which final-result-only checkers cannot localize.
Overall, these limitations motivate a state-based and scalable evaluation paradigm that assesses agents in realistic scenarios. 

To address these limitations, this paper proposes \textbf{\textit{STAGE-Claw}} 
(\textbf{\textit{S}}tate-based, \textbf{\textit{T}}ool-integrated, 
\textbf{\textit{A}}gent task \textbf{\textit{G}}eneration and 
\textbf{\textit{E}}valuation), an automated framework for constructing and evaluating agent benchmarks in realistic environments. 
STAGE-Claw addresses these challenges from three aspects.
\textbf{State-based evaluation} verifies whether an agent's actions produce the expected state changes in the environment, rather than merely checking outputs or artifacts. 
\textbf{Automated construction} and \textbf{Process-aware diagnosis} refer to evaluating agents by automatically generating and validating benchmark instances and analyzing fine-grained metrics to localize task failure reasons. 
As shown in Figure \ref{fig:main}, STAGE-Claw automatically creates task instances from the task hint word, validates their verifiability, difficulty, and reproducibility, executes target agents in the generated environments, and evaluates their performance by verifying persistent system-state changes across tools. 
This design shifts the evaluation from final-artifact checking to state-based assessment of agent behavior in realistic scenarios. 
We build 40 challenging tasks and conduct a detailed analysis of 11 frontier models' test results. 
Overall, our contributions are summarized as follows:

\begin{itemize}
    \item  This paper proposes STAGE-Claw, a framework that systematically automates the construction and validation of state-based agent evaluation instances in realistic scenarios.

    \item Using STAGE-Claw, we build a state-based benchmark of 40 challenging tasks grounded in 5-group realistic scenarios, covering workflows that require cross-source reasoning, tool state updates, and cross-tool consistency.

    \item We evaluate 11 frontier models on STAGE-Claw and conduct a detailed analysis of the task trajectory and results, which contribute insights for developing reliable, state-based, and extensible agent evaluation systems.
    
\end{itemize}

\section{STAGE-Claw}

In this section, we introduce \textbf{STAGE-Claw}, a four-stage automated framework for constructing and evaluating state-based agent benchmarks. 

\paragraph{Formalized Definition.} Each benchmark instance is formulated as a state-transformation problem over a reconstructable real environment:
\[
\mathcal{B}=(q,E_0,G,R,V),
\]
where \(q\) is a task prompt, \(E_0\) is an initial tool environment, \(G\) specifies a target final state, \(R\) is a scoring rubric, and \(V\) is an executable verifier. An agent observes only \((q,E_0)\) at the beginning, while \((G,R,V)\) is reserved for evaluation to prevent information leakage. Given an agent policy \(\pi\), execution from the initial state \(\boldsymbol{s}_0\) produces a trajectory
\[
\tau_{\pi}=(\boldsymbol{s}_0,o_0,a_0,\ldots,a_{H-1},o_H,\boldsymbol{s}_H),
\]
where \(o_t\) is the environment observation, \(a_t\) is the agent's action for tool use, and \(\boldsymbol{s}_H\) is the final environment state. The agent succeeds not by merely producing a final textual answer, but by transforming \(E_0\) into a final state \(G\) where a \(V\) is utilized to evaluate its correctness.

\paragraph{\textbf{Implementation Environment.}}
STAGE-Claw evaluates agents in real computing environments, where user requests often span multiple applications. 
The environment can be regarded as a collection of all the states of the tools. Initial environment $E_0=(\mathcal{T},\boldsymbol{s}_0)$, where $\mathcal{T}$ denotes the set of available tools and $\boldsymbol{s}_0$ denotes their initial joint state. 
The global environment state at step $t$ is represented as $\boldsymbol{s}_t=\{s_t^\tau\}_{\tau\in\mathcal{T}}$, where $s_t^\tau$ is the state of tool $\tau$. 
Agents receive user-level access to realistic tools, including the file system, browser, terminal, calendar, email, reminders, and notes, which support read, write, and execution operations over persistent application states. 
Tasks are modeled as transitions from the initial state $\boldsymbol{s}_0$ to an acceptable final-state set $\mathcal{S}_G$ specified by a target $G$, and are considered successful when $\boldsymbol{s}_H\in\mathcal{S}_G$, indicating that the agent has produced the required persistent changes while preserving relevant existing state.

\begin{figure}
  \includegraphics[width=0.45\textwidth]{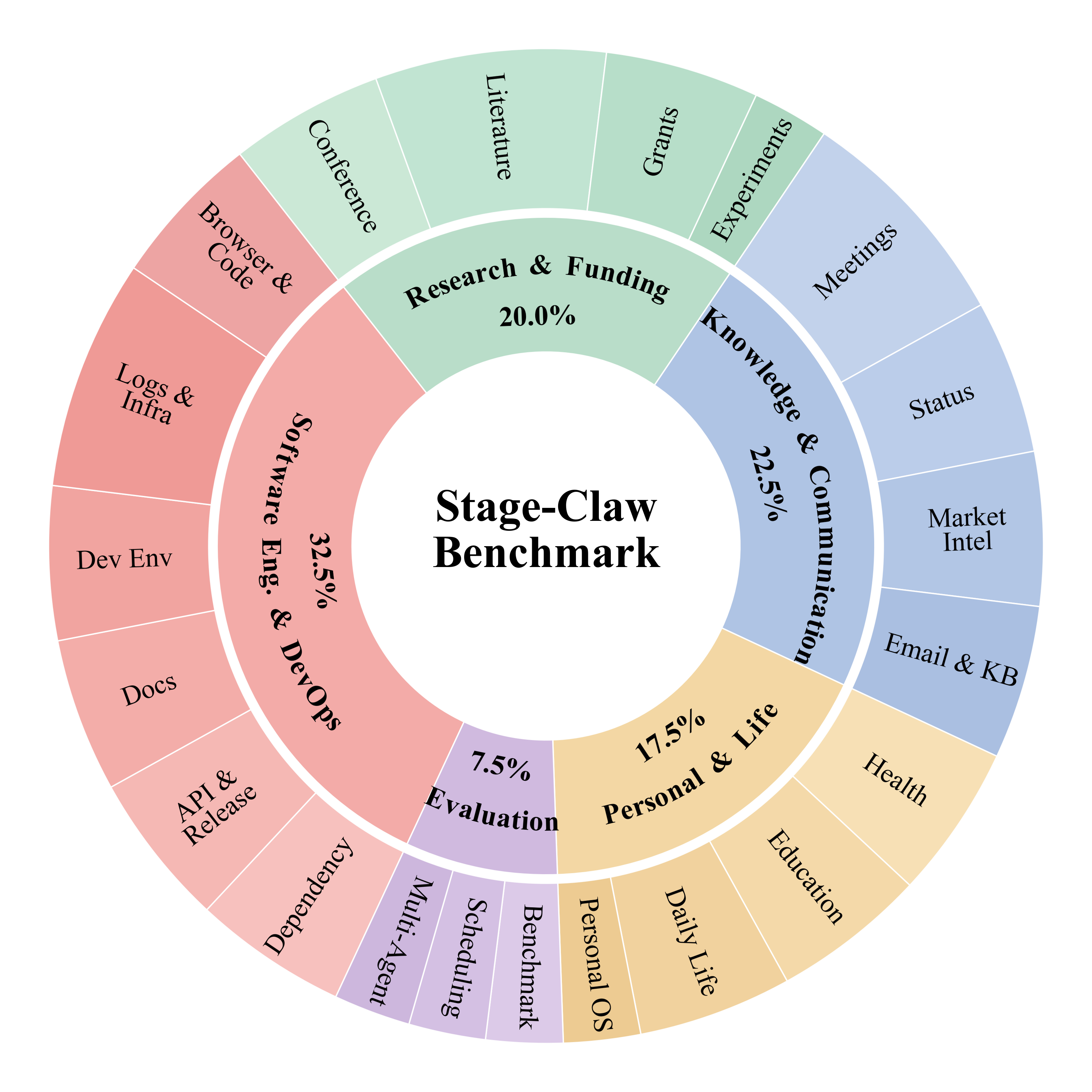}
  \vspace{-10pt}
  \caption{Statistics of task categories.}
  \label{fig:task_category_pie}
\vspace{-10pt}
\end{figure}

\subsection{Stage 1: Benchmark Authoring}
\label{sec:benchmark_authoring}

The first stage automatically constructs executable benchmark instances from task hint words. 
We manually curate 40 real assistant scenarios as task hint words. 
For each task hint, STAGE-Claw invokes a benchmark-authoring agent to instantiate an executable task.
The authoring agent acts only as a benchmark designer. 
Explore realistic user needs and imagine several challenging scenarios, then select a scenario with sufficient complexity.  
Based on the selected scenario, the authoring agent constructs a task instance containing a task prompt, an environment-construction guide, the corresponding ground truth, and an executable verification program aligned with the ground truth.

To ensure sufficient complexity, each task is designed to be multi-step, involve multiple tools or information sources, be reconstructable from a clean state, and support objective state-based evaluation. We also deliberately add some operators with different difficulty types (illustrated in Table \ref{tab:difficulty_mechanisms} in Appendix \ref{sec:Difficulty}) to model the complexity of real-world scenario tasks.
In general, we build 40 tasks into 5 groups. Figure \ref{fig:task_category_pie} shows the task distribution.

\subsection{Stage 2: Benchmark Validation}
\label{sec:benchmark_validation}

Before admission into the evaluation set, each task instance is checked by an independent validation agent that does not solve or modify the task. The checker verifies four properties. 
\textit{Structure}, ensuring that each task includes environment construct guidance, visible task prompt, hidden ground truth, and the corresponding executable verify program.
\textit{Reproducibility}, rebuilding the environment twice from a clean state and comparing snapshots across tool states such as files, calendars, reminders, notes, and email. Ensure that the task environment can be consistently reconfigured to the same initial state each time.
\textit{Verifiability}, assessing whether the scoring is objective and executable.
\textit{Difficulty calibration}, checking whether the difficulty level of the task includes the following types of difficulty. Such as cross-source conflicts, hidden dependencies, noisy data, entity alignment, tool state updates, and cross-tool consistency.

Each dimension receives a status of 
\textcolor{passgreen}{\textbf{\textsc{Pass}}}, 
\textcolor{failred}{\textbf{\textsc{Fail}}}, 
\textcolor{blockedgray}{\textbf{\textsc{Blocked}}}, or 
\textcolor{warnorange}{\textbf{\textsc{Warn}}}. These statuses are aggregated into a weighted 100-point checker score. Instances scoring above a threshold (80 in our experiments) are accepted. 
Failed instances are returned to \textit{Stage 1} with diagnostics for targeted repair and re-submitted until they pass or reach the maximum number of repair attempts.

\begin{table*}[t]
\centering
\small
\setlength{\tabcolsep}{7pt}
\renewcommand{\arraystretch}{1.12}
\begin{tabular}{lccccc}
\toprule
\textbf{Benchmark} & \textbf{State-based} & \textbf{Multi-tool} & \textbf{Auditable} & \textbf{Auto Construction} & \textbf{Perturbation} \\
\midrule
AgentBench~\citep{liu2024agentbench}        & \ymark & \ymark & \ymark & \rmark & \rmark \\
GAIA~\citep{mialon2024gaia}                 & \rmark & \ymark & \rmark & \rmark & \rmark \\
$\tau$-bench~\citep{yao2024tau}             & \cmark & \cmark & \cmark & \rmark & \rmark \\
WebArena~\citep{zhou2024webarena}           & \cmark & \ymark & \cmark & \ymark & \rmark \\
OSWorld~\citep{xie2024osworld}              & \cmark & \cmark & \cmark & \ymark & \rmark \\
ToolBench~\citep{qin2024toolllm}            & \rmark & \cmark & \ymark & \cmark & \rmark \\
Terminal-Bench~\citep{merrill2026terminal}  & \cmark & \ymark & \cmark & \rmark & \rmark \\
PinchBench~\citep{kilo2026pinchbench}       & \ymark & \ymark & \ymark & \rmark & \rmark \\
Claw-Eval~\citep{ye2026claw}                & \cmark & \cmark & \cmark & \ymark & \cmark \\
\midrule
\textbf{STAGE-Claw (Ours)}                  & \cmark & \cmark & \cmark & \cmark & \cmark \\
\bottomrule
\end{tabular}
\caption{
Comparison of agent evaluation benchmarks along STAGE-Claw.
\textbf{State-based} checks explicit environment states;
\textbf{Multi-tool} requires coordinated tool use;
\textbf{Auditable} supports traces, snapshots, rubrics checkers;
\textbf{Auto Construction} enables automatic or programmatic task construction;
and \textbf{Perturbation} introduces controlled noise, conflicts, or errors.
Green checks, yellow circles, and red crosses denote full, partial, and no core support.
}
\vspace{-10pt}
\label{tab:benchmark_comparison}
\end{table*}

\subsection{Stage 3: Agent Execution}

This stage involves testing the qualified tasks. For each validated benchmark, STAGE-Claw resets the initial state, reconstructs the environment from the environment-construction guide, and creates an isolated execution workspace containing only the visible task prompt, related documents, and tools. 
The evaluated agent must complete the task within a predefined time budget using only the available instructions and tools. 
During execution, it may interpret requirements, interact with tools and update tool states, inspect intermediate results, and produce final outputs. 
Once the agent finishes, STAGE-Claw records execution metadata, including run status, elapsed time, and tool-interaction traces. 
It also captures a snapshot of the final environment. 
These recorded states are sent to the evaluation stage, where the evaluator assesses whether the agent’s outputs and final-state record satisfy the task requirements.

\subsection{Stage 4: State-Based Evaluation}

The final stage performs a state-based evaluation after the evaluated agent completes the task execution. 
After execution, STAGE-Claw runs the task-specific verifier to check whether the recorded states align with the hidden ground truth $G$. 
The evaluator checks file output, tool-state updates, and formatting constraints, then returns a structured evaluation report. 
LLM-assisted adjudication is used only as a guarded fallback. 
It is triggered when executable verification fails to execute, times out, or cannot access a required snapshot field because of evaluator-side tool instability. 
All evaluation paths enforce a unified total-score format for automatic parsing and aggregation. 
Finally, STAGE-Claw records the agent score, execution status, and evaluation metadata and links them with execution logs to support detailed analysis.

Overall, we constructed 40 challenging tasks using STAGE-Claw. As summarized in Table \ref{tab:benchmark_comparison}, STAGE-Claw combines state-based evaluation, multi-tool workflows, auditable verification, automated construction, and controlled perturbation in one workflow. This design makes STAGE-Claw a scalable state-based evaluation framework rather than a collection of manually written agent tasks.

\begin{table*}[t]
\centering
\resizebox{\textwidth}{!}{
\begin{tabular}{ccccccc}
\toprule
\textbf{Model} & \textbf{Score} & \textbf{First-Pass} & \textbf{Time (s)} & \textbf{Tokens (M)} & \textbf{Cost (\$)} & \textbf{Tool Calls} \\
\midrule
\textbf{Claude-Opus-4.7} & \textbf{77.1} & \textbf{80.0\%} & 422.8 & 1.242 & \$6.55 & 40.6 \\
\textbf{Claude-Sonnet-4.6} & \underline{69.43} & 65.0\% & 733.8 & 1.265 & \$3.98 & 42.2 \\
\textbf{GPT-5.5} & 69.19 & 65.0\% & \underline{316.4} & \underline{0.405} & \$1.33 & 28.9 \\
\textbf{GPT-5.4} & 65.37 & 52.5\% & \textbf{192.5} & \textbf{0.236} & \$0.69 & \textbf{21.8} \\
\textbf{Gemini-3.1-Pro} & 65.5 & \underline{67.5\%} & 742.4 & 0.642 & \$2.65 & 52.6 \\
\textbf{DeepSeek-V4-Pro} & 59.78 & 60.0\% & 744.4 & 0.620 & \$0.28 & \underline{28.3} \\
\textbf{Doubao-Seed-2.0-Pro} & 61.07 & 50.0\% & 325.6 & 0.531 & \underline{\$0.27} & 28.4 \\
\textbf{GLM-5} & 67.40 & 60.0\% & 423.7 & 0.694 & \$0.73 & 40.5 \\
\textbf{Kimi-K2.6} & 59.05 & 55.0\% & 641.3 & 0.921 & \$0.93 & 54.9 \\
\textbf{MiniMax-M2.7} & 47.53 & 35.0\% & 937.5 & 0.529 & \textbf{\$0.17} & 34.8 \\
\textbf{Qwen3.5-Plus} & 59.74 & 57.5\% & 406.2 & 0.882 & \$0.30 & 59.3 \\
\bottomrule
\end{tabular}
}
\caption{
Performance and efficiency comparison of large language models on STAGE-Claw. Score is the average valid run task score. First-Pass denotes the proportion of tasks scoring above 60 in the first round. Time, Tokens, Cost, and Tool Calls report average run time, token usage, API cost, and tool invocations of each task.
Bold values indicate the best result for each metric, and underlined values indicate the second-best result.
}
\label{tab:model_performance}
\vspace{-5pt}
\end{table*}

\section{Evaluation}


\subsection{Task Construction}
\label{sec:benchmark_construction}

\paragraph{Construction Settings.}  Benchmark authoring agent is instantiated with Claude-Sonnet-4.6 \cite{anthropic2026claudesonnet46}.
The full prompts are shown in Appendix \ref{app:task_prompt}. 
Each candidate task is checked by the checker agent (test results exceeded 80 points), which requires an average of 2.67 repair iterations. 
All accepted tasks can be verified by executable verification programs without the need for LLM-assisted adjudication. 
Finally, human annotators check each accepted task for scenario realism, task completeness, instruction clarity, ground-truth correctness, and evaluator-rubric alignment. The annotation details are provided in Appendix~\ref{app:human_annotation}. All task construction, repair, and checker-based validation are fully automated, human involvement is only used for quality review.

\paragraph{Construction Budget and Scope.} 
Each STAGE-Claw task requires automated construction, verification, evaluation, and human audit. 
Building an accepted task costs about 35 to 40\$ and takes 1-2 hours, excluding model-evaluation runs. 
The time cost to each task for each model is 3 to 15 minutes, and the API cost is 0.17 to 6.55\$.
Considering the substantial time and API costs of constructing and evaluating tasks, we build a set of 40 challenging tasks and position it as a high-quality pilot benchmark rather than an exhaustive coverage of all personal-agent scenarios. 
We evaluate each model on each task once in our experiments and include a limited repeated-run diagnostic in Appendix~\ref{app:Repeated-run}.
Our goal is to demonstrate the feasibility and diagnostic value of automated state-based evaluation in realistic environments.

\subsection{Evaluated Models}

We evaluated 11 state-of-the-art models spanning multiple model families: Claude-Opus-4.7 \cite{anthropic2026claudeopus47},
Claude-Sonnet-4.6 \cite{anthropic2026claudesonnet46}, 
DeepSeek-V4-Pro \cite{deepseek2026v4pro}, 
Qwen3.5-Plus \cite{alibabacloud2026qwen35plus},
GPT-5.5 \cite{openai2026gpt55},
GPT-5.4 \cite{openai2026gpt54}, 
Gemini-3.1-pro-preview \cite{google2026gemini31propreview}, 
Doubao-Seed-2.0-Pro \cite{bytedance2026seed20pro}, 
GLM-5 \cite{glm5team2026glm5}, 
Kimi-k2.6 \cite{moonshot2026kimik26}, 
and MiniMax-M2.7 \cite{minimax2026m27}. 
All models were assessed in the 40 benchmark tasks constructed for this study.

\subsection{Evaluation Settings}

All models were tested within the framework of OpenClaw, with reasoning disabled and temperature set to 0 when configurable (otherwise using the provider-fixed temperature).
This setting minimizes sampling variability and improves reproducibility across runs. 
Each benchmark task was executed in an isolated operating-system environment, and a freshly initialized OpenClaw agent was instantiated for each run to avoid interference from persistent memory, cached state, or prior context. 

Scores in Table \ref{tab:model_performance} show each model’s first valid run per task. 
A valid run requires a correctly reconstructed environment and a completed evaluation. 
Given that tasks involve interactions with real tools, including calendar events, emails, and reminders, executions were performed serially to avoid conflicts between concurrent tasks. This setup ensures that each trial reflects the agent’s ability to complete the task based solely on the visible prompt and accessible tool interfaces, providing a controlled and reproducible evaluation environment. 

\subsection{Main Results}
\label{sec:main_results}

Table~\ref{tab:model_performance} reports the performance and efficiency of different LLMs in STAGE-Claw under the OpenClaw framework. Overall, the results reveal a clear trade-off between task performance and execution efficiency.

\textbf{Model performance and execution efficiency exhibit a clear trade-off.}
Claude-Opus-4.7 achieves the strongest overall performance, with the highest average score of 77.1 and the highest First-Pass rate of 80.0\%, demonstrating the highest reliability among all evaluated models. Claude-Sonnet-4.6 obtains the second-best average score, while Gemini-3.1-Pro achieves the second-best First-Pass rate. However, these strong-performing models generally require higher computational overhead, such as longer execution time, larger token consumption, or higher API cost. In contrast, GPT-5.4 shows the best efficiency profile, achieving the lowest latency, the lowest token usage, and the fewest tool calls while maintaining a competitive score. GPT-5.5 further provides a strong balance between performance and efficiency, approaching Claude-Sonnet-4.6 in score while using substantially fewer resources.

\textbf{Tool-use frequency does not directly determine task performance.}
Models such as Qwen3.5-Plus (59.3) and Kimi-K2.6 invoke tools (54.9) far more frequently than GPT-5.4 (21.8), yet obtain lower average scores. Conversely, GPT-5.4 completes tasks with the fewest tool calls while outperforming several models with heavier tool usage. This suggests that effective tool use depends more on planning quality, tool selection, and result verification than on the raw number of tool invocations. Excessive tool calls may therefore indicate inefficient coordination rather than stronger task-solving ability.

\begin{figure}
  \includegraphics[width=0.5\textwidth]{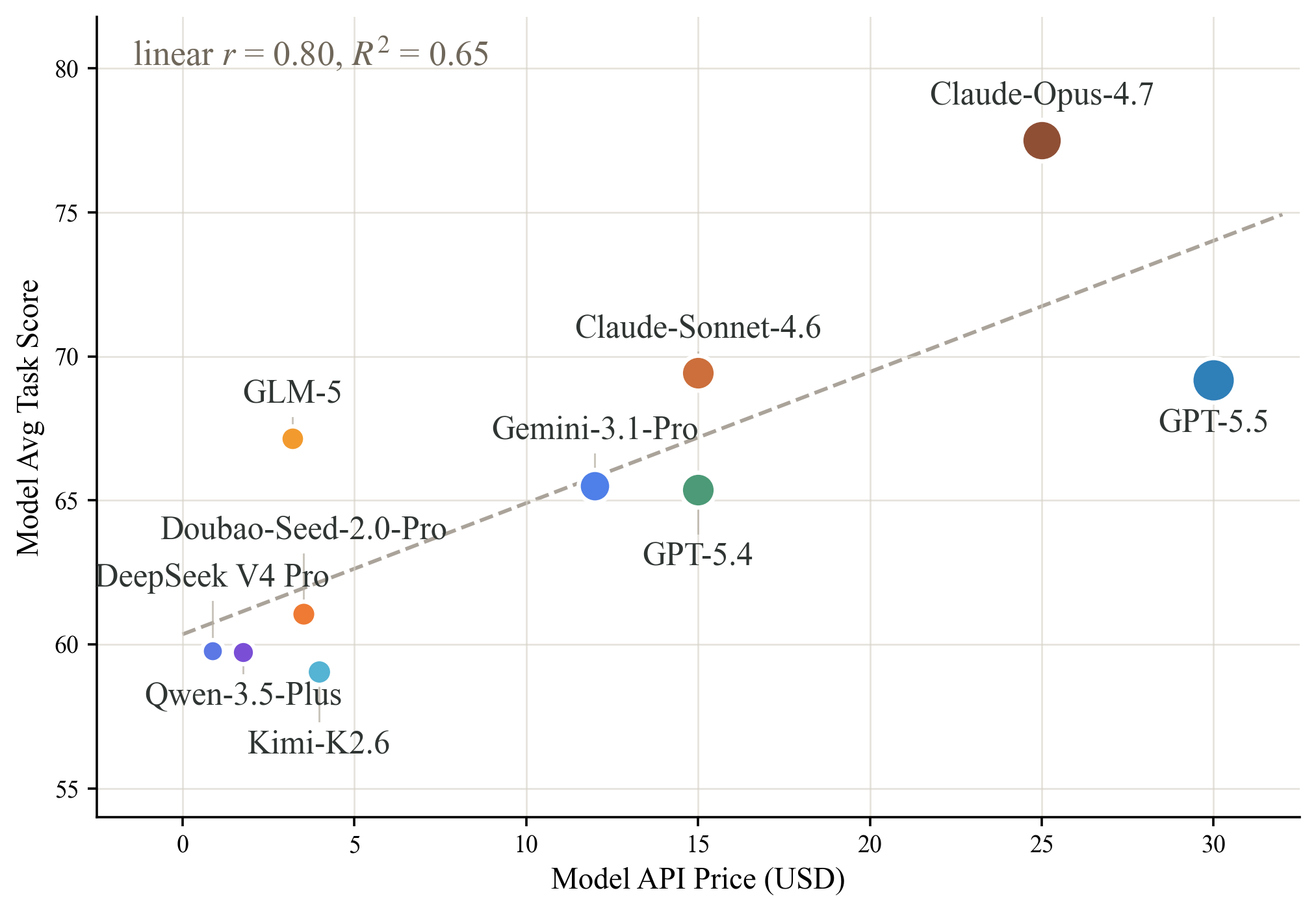}
  \caption{Performance comparison between the unit model api price, and the task score. The x-axis represents the unit price of the API, and the y-axis represents the average model task score.}
  \label{fig:score_unit}
\vspace{-10pt}
\end{figure}

\textbf{Higher API unit price is positively correlated with stronger task performance, but does not fully explain model quality.}
As shown in Figure \ref{fig:score_unit}, model average task score exhibits a strong positive linear correlation with model API price. This suggests that more expensive models tend to achieve better task performance on STAGE-Claw. Claude-Opus-4.7 is a representative example, combining a high API price with the highest average score. However, the correlation is not absolute. GLM-5 achieves a relatively high score at a low API price, while GPT-5.5 is more expensive but does not outperform Claude-Opus-4.7 or Claude-Sonnet-4.6. 
We speculate that GPT-5.5's relative underperformance may partly stem from its configuration with reasoning (thinking) modules turned off, which prioritizes execution efficiency over deliberative planning and multi-step reasoning. 
These observations indicate that price is an important but insufficient proxy for agent capability and that cost-effectiveness varies substantially across models.

\begin{figure*}[t]
  \centering
  \begin{minipage}{0.49\textwidth}
    \centering
    \includegraphics[width=\linewidth]{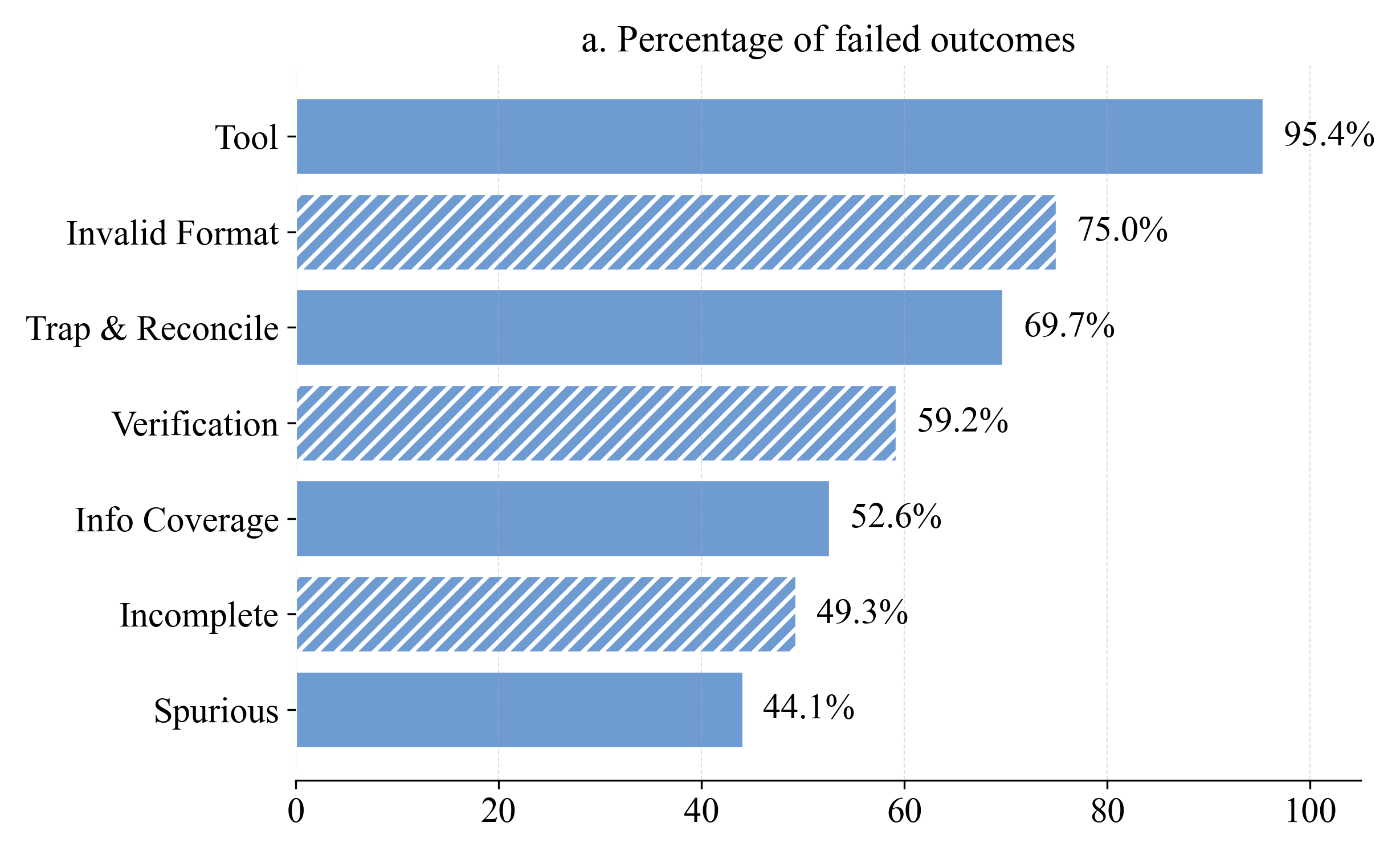}
  \end{minipage}
  \hfill
  \begin{minipage}{0.49\textwidth}
    \centering
    \includegraphics[width=\linewidth]{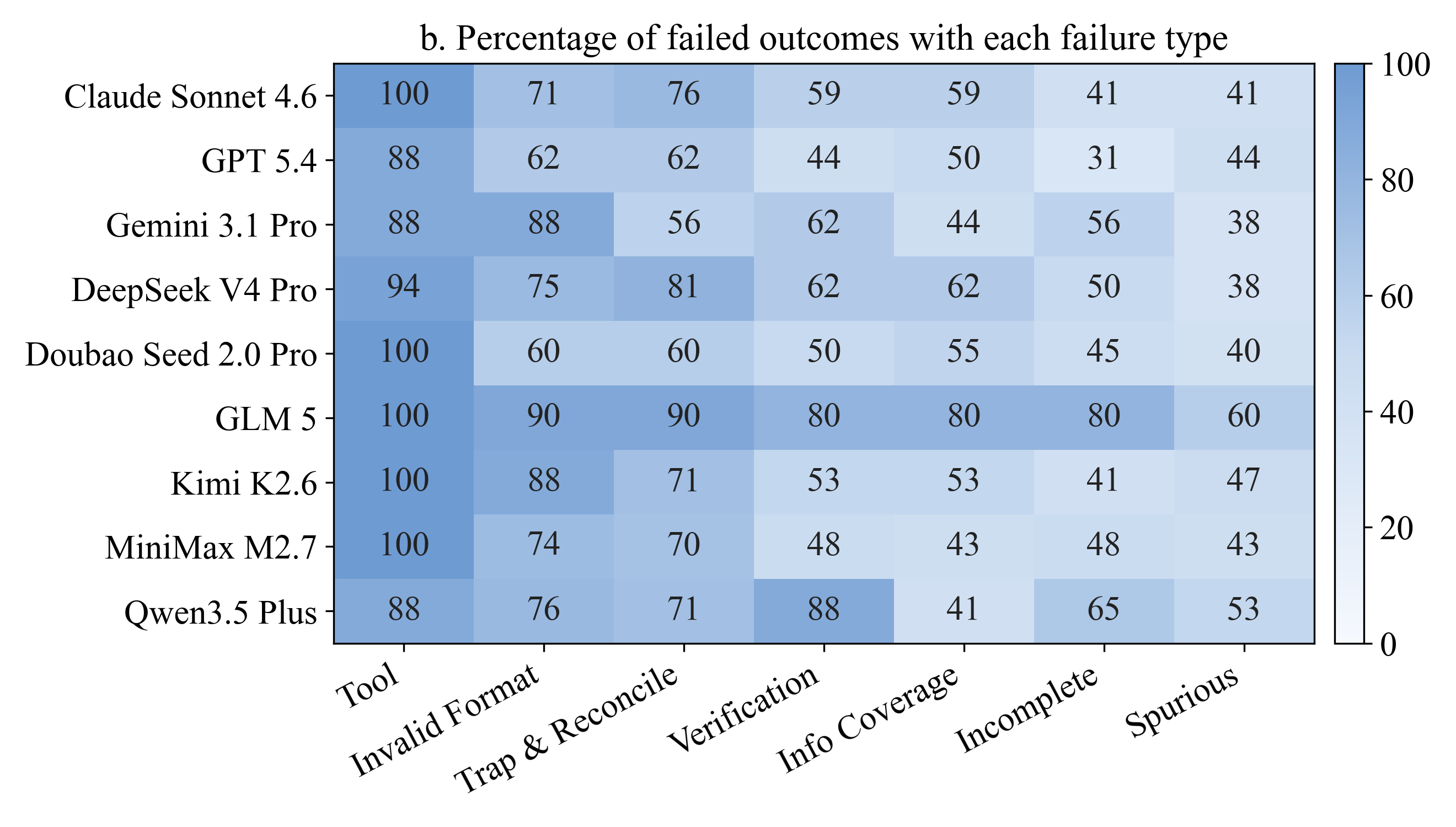}
  \end{minipage}
  \begin{minipage}{1\textwidth}
    \centering
    \includegraphics[width=\linewidth]{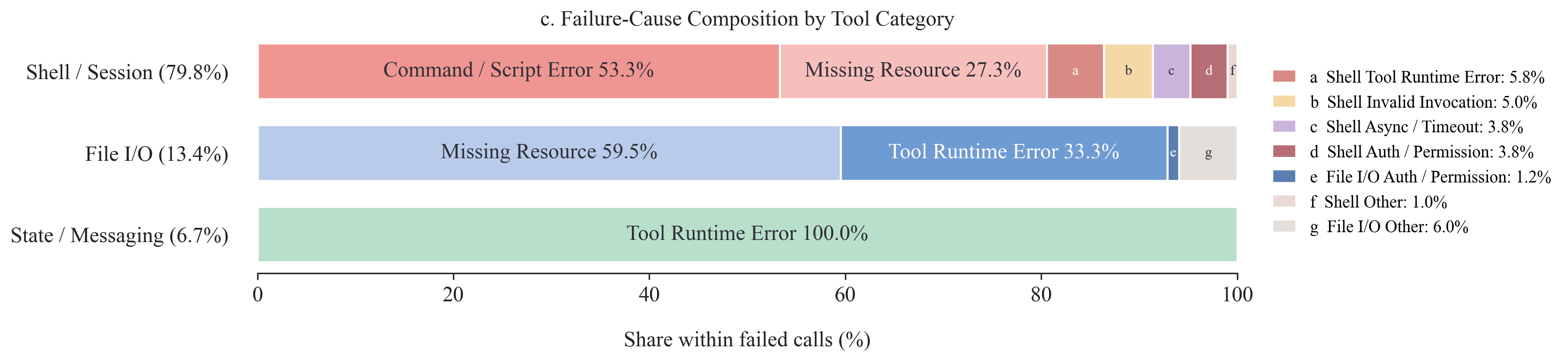}
  \end{minipage}
  \vspace{-5pt}
  \caption{Failure-type analysis over non-passing outcomes. Categories are multi-label, so percentages do not sum to 100. \textbf{\textit{a.}} prevalence of each failure type among failed outcomes. \textbf{\textit{b.}} model-wise percentage of failed outcomes containing each type. \textbf{\textit{c.}} proportion of tool usage and the error types for the corresponding tools}
  \label{fig:failure_analysis}
  \vspace{-5pt}
\end{figure*}

These findings show that STAGE-Claw evaluates not only final task success, but also how reliably and efficiently models operate in realistic scenarios. 
The results highlight the need to jointly consider task performance, latency, realized run cost, token usage, API unit price, and tool-use behavior when assessing LLM agents.

\section{Analysis}

\subsection{Failure Analysis}
\label{sec:failure_analysis}

We further analyze non-passing runs to characterize common failure patterns. Since a single run may exhibit multiple errors, our analysis is multi-label rather than mutually exclusive. 
Figure \ref{fig:failure_analysis}.a summarizes the resulting distribution.

The most prevalent failure mode is \textit{\textbf{Tool Failure}}, appearing in 95.4\% of non-passing runs. These failures involve missing, incomplete, or incorrectly routed writes to real tools such as calendar, notes, reminders, or email, indicating that agents often generate plausible intermediate outputs but fail to correctly update the underlying environment. 
The second most frequent category is \textit{\textbf{Invalid Format}} (75.0\%), including incorrect JSON format, missing required fields, incorrect filenames, or verification reports without required metadata. 
\textit{\textbf{Trap and Reconciliation Failures}} (69.7\%) and \textit{\textbf{Verification Failures}} (59.2\%) further show that agents struggle with source-of-truth selection, entity alignment, timezone and deadline reasoning, deduplication, noise filtering, and evidence-backed validation. 
\textbf{\textit{Information Coverage Errors}} (52.6\%), \textbf{\textit{Incomplete Execution}} (49.3\%), and \textbf{\textit{Spurious Outputs}} (44.1\%) are also common, suggesting that current agents have difficulty integrating reliable extraction, verification, and conservative state updates into a coherent workflow.
Figure \ref{fig:failure_analysis}.b shows the percentages in model of failed outcomes.

\subsection{Tool-Use Analysis}
\label{sec:tool_analysis}

In this section, we analyze  the frequency of tool calls and the errors that occur in the frequent tool calls.
Figure \ref{fig:failure_analysis}.c reports both the tool call usage distribution and the failure cause composition, where failure causes are normalized over failed tool calls within each category. 
Among all categorized tool calls, Shell/Session tools are used most frequently, accounting for 79.8\% of calls, followed by File I/O tools at 13.4\%. 
State/Messaging tools are invoked much less often, representing only 6.7\% of calls.

Despite their different usage frequencies, the failure patterns vary substantially across tool categories. 
Shell/Session failures are dominated by command or script errors (53.3\%), followed by missing resources or paths (27.3\%).
This suggests that shell-based interaction primarily fails due to incorrect command construction, missing dependencies, or invalid assumptions about the runtime environment. 
File I/O failures exhibit a distinct pattern: missing resources or paths account for 59.5\% of failed calls, while tool runtime errors account for 33.3\%. 
This indicates that accurate resource localization and path validation are critical for reliable file operations. 
By contrast, all observed State/Messaging failures are attributed to tool runtime errors, suggesting that operations involving persistent application state or external communication channels remain particularly brittle.

Overall, tool failures are category-specific rather than driven by a single generic error mode. 
Shell/Session tools require more robust command generation and dependency checking. File I/O tools depend heavily on reliable resource validation, and State/Messaging tools require stronger recovery mechanisms for persistent-state updates.

\subsection{Effectiveness of State-based Evaluation}

To examine whether state-based evaluation can better capture unexpected failures in realistic scenarios, we construct a virtual-state variant of STAGE-Claw. 
In this setting, each task is modified to replace actual tool-state changes with textual outputs that simulate the expected state updates. 
We run this virtual setting on the same tasks using two representative models, DeepSeek-V4-Pro and Qwen3.5-Plus, and compare the results with the original state-based evaluation.

\begin{table}[t]
\centering
\small
\begin{tabular}{p{2.35cm}ccc}
\toprule
\textbf{Model} & State-based & Virtual & $\Delta$ \\
\midrule
DeepSeek-V4-Pro & 59.78 & 66.71 & +6.93 \\
Qwen3.5-Plus & 59.74 & 64.57 & +4.83 \\
\midrule
\textbf{Cause category} & \#Tasks & Proportion & Avg. $\Delta$ \\
\midrule
Execution failure & 7 & 8.75\% & +76.57 \\
Real-state gap & 13 & 16.25\% & +14.05 \\
Other errors & 3 & 3.75\% & -2.67 \\
\bottomrule
\end{tabular}
\caption{The upper section shows a comparison between state-based and virtual evaluation, $\Delta$ denotes Virtual - State-based. The lower section breaks down the causes of these gaps.}
\vspace{-10pt}
\label{tab:state_vs_virtual}
\end{table}

As shown in Table~\ref{tab:state_vs_virtual}, the virtual setting yields higher scores than the state-based setting for both models. 
However, this gap does not reflect better task completion. 
Instead, it suggests that output-only evaluation can overestimate agent performance by ignoring whether the agent actually modifies persistent tool states. 
In real tool-integrated environments, agents may fail due to execution failure and real-state gap, such as invalid state writes, missing generated artifacts, and tool-side effects. 
These failures are naturally exposed by state-based evaluation, but can be hidden when the task is reduced to textual state simulation. 
Therefore, the comparison indicates that the state-based evaluation provides a more faithful assessment of agent capability in realistic tool-use scenarios.

\subsection{Memory Perturbation Analysis}
\label{sec:memory_perturbation}

\begin{table}[t]
\centering
\small
\setlength{\tabcolsep}{3.5pt}
\renewcommand{\arraystretch}{0.9}
\resizebox{\linewidth}{!}{
\begin{tabular}{lcccc}
\toprule
\textbf{Task} & \textbf{Base} & \textbf{Noise} & \textbf{Misleading} & \textbf{Conflict} \\
\midrule
AI Conference Tracker & 77.0 & $23.0_{-54.0}$ & $23.0_{-54.0}$ & $23.0_{-54.0}$ \\
Benchmark Administrator & 85.0 & $56.0_{-29.0}$ & $35.0_{-50.0}$ & $49.0_{-36.0}$ \\
Competitor Radar & 62.0 & $57.0_{-5.0}$ & $55.0_{-7.0}$ & $55.0_{-7.0}$ \\
Email Message Triage Center & 45.0 & $40.0_{-5.0}$ & $22.0_{-23.0}$ & $22.0_{-23.0}$ \\
\midrule
\textbf{Average} & \textbf{67.3} & $\mathbf{44.0}_{\mathbf{-23.3}}$ & $\mathbf{33.8}_{\mathbf{-33.5}}$ & $\mathbf{37.3}_{\mathbf{-30.0}}$ \\
\bottomrule
\end{tabular}
}
\caption{Qwen3.5-Plus scores under manually constructed memory perturbations on four STAGE-Claw tasks. Subscripts indicate score drops relative to the Base setting.}
\label{tab:memory_perturbation}
\vspace{-8pt}
\end{table}

Persistent memory can also affect state-based agent execution, especially when it contains irrelevant, stale, or conflicting information. 
We therefore conduct a small diagnostic study on four STAGE-Claw tasks using Qwen3.5-Plus. 
For each task, we move key task-specific requirements from the visible instruction into the agent's persistent memory and lightly revise the task prompt so that the task remains natural and memory-dependent. 
We then manually construct three perturbation types: \textit{Noise}, which adds irrelevant memory entries around the key facts, \textit{Misleading}, which injects plausible but incorrect rules or values and \textit{Conflict}, which introduces inconsistent versions of the same key facts.

Table~\ref{tab:memory_perturbation} reports the resulting scores. 
All three perturbation types reduce performance relative to the base memory-dependent setting, with \textit{Misleading} and \textit{Conflict} causing larger drops in our sample. 
These results are intended as diagnostic evidence rather than a comprehensive benchmark-wide conclusion. 
They suggest that memory quality can influence both agent decisions and downstream state updates, motivating future evaluation protocols and agent mechanisms that track memory provenance, recency, and conflicts in addition to tool-execution reliability.

\section{Related Work}

\paragraph{Agent Benchmarks for Tool Use and Interactive Environments.}
Recent agent benchmarks have expanded LLM evaluation beyond isolated text generation toward tool use, multi-step reasoning, and interactive task completion.
General-purpose benchmarks such as AgentBench~\cite{liu2024agentbench} and GAIA~\cite{mialon2024gaia} evaluate planning, reasoning, web access, and external tool use, while function-calling benchmarks such as BFCL~\cite{patil2025berkeley} focus on whether models can correctly select, parameterize, and invoke tools.
Another line of work evaluates agents in interactive environments.
Mind2Web~\cite{deng2023mind2web}, WebArena~\cite{zhou2024webarena}, VisualWebArena~\cite{koh2024visualwebarena}, and WorkArena~\cite{drouin2024workarena} study web and enterprise workflows, and OSWorld~\cite{xie2024osworld} evaluates agents in real operating-system and desktop applications.
AppWorld~\cite{trivedi2024appworld} and \(\tau\)-bench~\cite{yao2024tau} further emphasize controllable app or tool-agent-user interactions.
These benchmarks provide progress in realistic interaction, but they are primarily designed around fixed tasks, simulated APIs, GUI control, or transactional tool use.

\paragraph{Personal-Agent Benchmarks and State-based Evaluation.}
Recent benchmarks have moved closer to OpenClaw-oriented personal agents, where LLMs interact with tools, memory, execution environments, and control logic~\cite{openclaw2026,li2026clawsbench}.
PinchBench~\cite{kilo2026pinchbench}, WildClawBench~\cite{WildClawBench}, ClawsBench~\cite{li2026clawsbench}, and Claw-Eval~\cite{ye2026claw} evaluate practical workflows such as scheduling, email, research, file management, and multi-turn tool use.
Many existing agent benchmarks rely on manually curated tasks, sandboxed artifacts, simulated workspaces, or coarse rubrics, limiting their scalability and ability to diagnose persistent-state errors. 
STAGE-Claw addresses these limitations by automatically generating benchmark instances from task hint words, instantiating them in real-application environments, and evaluating agents via snapshots of the final system state.

\section*{Conclusion}

This paper introduces STAGE-Claw, an automated state-based framework for constructing and evaluating realistic scenario benchmarks. 
By assessing state-based snapshots rather than textual responses, STAGE-Claw enables more faithful evaluation of agent ability in multi-tool workflows. 
In addition, the automated pipeline provides a practical path toward scaling realistic scenario agent benchmarks while preserving reproducibility and auditable evaluation. 
Experiments on 40 tasks show that current agents still struggle to complete complex tasks that involve multiple tool invocations in real scenarios. 
These results highlight the need for state-based evaluation and provide practical insights for building more reliable assistant agents.

\section*{Limitations}
STAGE-Claw is designed for realistic state-based evaluation, but this design also brings limitations. 
First, the benchmark currently contains 40 accepted tasks, and the main evaluation follows a first-valid-run protocol with only limited repeated-run diagnostics. 
Thus, the results should be viewed as a controlled snapshot rather than a statistically exhaustive leaderboard. 
Second, constructing high-quality state-based tasks remains resource-intensive. 
Although task authoring, repair, and checker-based validation are automated after human-provided task hints, each accepted task still requires realistic scenario design, deterministic environment setup, executable verifier implementation, and quality checking, which together incur nontrivial time and API costs. 
Third, evaluation in real personal-computing environments means that scores may reflect not only model capability but also OpenClaw, tool wrappers, OS configuration, permissions, and runtime stability. 
This improves realism but may introduce platform-dependent failures.



\bibliography{custom}



\appendix
\section{Appendix}

\subsection{Evaluation Separation and Blindness}

To preserve evaluation validity, the benchmark construction protocol enforces a strict separation between the agent-facing task prompt and the evaluator-only materials. The agent receives only the question prompt and the initialized environment specified by the environment-construction document. In contrast, the evaluator has access to ground truth, which may contain gold answers, scoring scripts, structured checklists, or other verification procedures.

The agent-facing prompt must not include any of the following: gold labels, expected final answers, hidden constraints, scoring rubrics, evaluator scripts, checksum values, or explicit references to the verification logic. This constraint ensures that task success reflects the agent's ability to operate in the environment rather than its ability to exploit leaked evaluation information.

\subsection{Design Criteria}

Each task produced under this protocol is required to satisfy the following criteria.

\begin{enumerate}
    \item \textbf{Multi-step execution.}
    The task must require at least three distinct operations, such as locating files, extracting information, transforming data, writing an output file, or interacting with an application.

    \item \textbf{Cross-tool grounding.}
    The task must require at least two different tools or information sources. Examples include filesystem access and terminal commands, browser interaction and document parsing, or local data inspection and application state modification.

    \item \textbf{Reproducibility.}
    The environment construction procedure must produce the same initial state on every run. All seeded files, directories, and system configurations should be specified explicitly.

    \item \textbf{Blind task execution.}
    The task prompt shown to the agent must not contain oracle information. The agent should only know the user-facing objective, the relevant working directory, and operational constraints.

    \item \textbf{Objective gradability.}
    The final result must be verifiable without subjective human judgment. Acceptable evaluation methods include deterministic scripts, exact-match checks, structured rubrics with binary criteria, or reproducible checklist-based validation.

    \item \textbf{Time independence.}
    The task must not depend on real clocks, real-time APIs, or mutable external states unless those states are explicitly mocked or seeded during environment construction.
\end{enumerate}

\subsection{Difficulty Categories}
\label{sec:Difficulty}

In order to simulate the difficulty and complexity of the task in a real situation, we have carried out complex processing on the task. The specific types of processing are shown in Table \ref{tab:difficulty_mechanisms}.

\begin{table}[h]
\centering
\small
\begin{tabularx}{\linewidth}{>{\centering\arraybackslash}m{2.1cm}|X}
\toprule
\textbf{Category} & \textbf{Difficulty Mechanism} \\
\midrule
\makecell{\textbf{Data}\\\textbf{Inconsistency}} & Cross-source conflicts, Heterogeneous timestamp formats, Inconsistent status or Field values \\
\midrule
\makecell{\textbf{Dependency}\\\textbf{Reasoning}} & Hidden dependencies, Topological ordering, Prerequisite Implementation requirements \\
\midrule
\makecell{\textbf{Noise}\\\textbf{Filtering}} & Test records, Drafts, Deprecated items, Dead URLs, Empty or invalid entries \\
\midrule
\makecell{\textbf{Entity}\\\textbf{Format}} & Aliases, Email mappings, Abbreviations, Canonical IDs, Category mapping \\
\midrule
\makecell{\textbf{Output}\\\textbf{Precision}} & Strict field schemas, Deterministic sorting, Casing constraints, Date formats, Exact-match error messages \\
\bottomrule
\end{tabularx}
\caption{Difficulty categories and included mechanisms}
\label{tab:difficulty_mechanisms}
\end{table}

\subsection{Snapshot Tools}

\begin{table*}[t]
    \centering
    \small
    \renewcommand{\arraystretch}{1.25}
    \begin{tabularx}{\linewidth}{p{0.22\linewidth} X X}
        \toprule
        \textbf{Environment Type}
        & \textbf{Snapshot Method}
        & \textbf{Recorded State} \\
        \midrule

        File system
        & Traverse the target environment directory using
        find XXX\_Env/ -type f, sort the file list, and compute
        hashes for both file paths and file contents.
        & A deterministic digest of the file hierarchy and all task-related
        file contents. \\

        Calendar events
        & Query calendar events in a fixed future window using
        icalbuddy eventsFrom:today to:+30d, and compute a hash of
        the normalized output.
        & A snapshot of scheduled calendar events within the evaluation
        window. \\

        Reminders
        & Use AppleScript to query reminder names, e.g., osascript -e 'tell app "Reminders" to get name of reminders'.
        & The set of reminder entries visible to the system at snapshot time. \\

        Notes
        & Use AppleScript to query note titles, e.g., osascript -e 'tell app "Notes" to get name of notes'.
        & The set of note records relevant to the initialized environment. \\

        Email state
        & Use AppleScript or mail-client queries to count messages under
        task-specific labels, including drafts and flagged messages.
        & The number and status of task-relevant email records, such as drafts,
        labels, or flags. \\

        \bottomrule
    \end{tabularx}
    \caption{Snapshot methods for validating reproducible benchmark
    environments. Each snapshot records a deterministic representation of the
    corresponding environment state and can be compared before and after task
    execution.}
    \label{tab:environment_snapshot_methods}
\end{table*}

\subsection{Self-Review Checklist}

Before a benchmark task is finalized, the author performs a self-review using the following checklist.

\begin{enumerate}
    \item The environment builder can follow environment make documents without requiring additional clarification.
    \item The setup procedure is deterministic and reproducible.
    \item The agent-facing question prompt contains no ground-truth information.
    \item The evaluator can score the task using only the final state and the files in ground truth documents.
    \item The success condition is singular, explicit, and unambiguous.
    \item Potential shortcuts or cheating strategies have been identified and blocked.
    \item The task admits a plausible harder variant for future benchmark scaling.
\end{enumerate}

\renewcommand{\tabularxcolumn}[1]{m{#1}}

\subsection{Repeated-run diagnostic.} 
\label{app:Repeated-run}

\begin{figure*}
  \includegraphics[width=\textwidth]{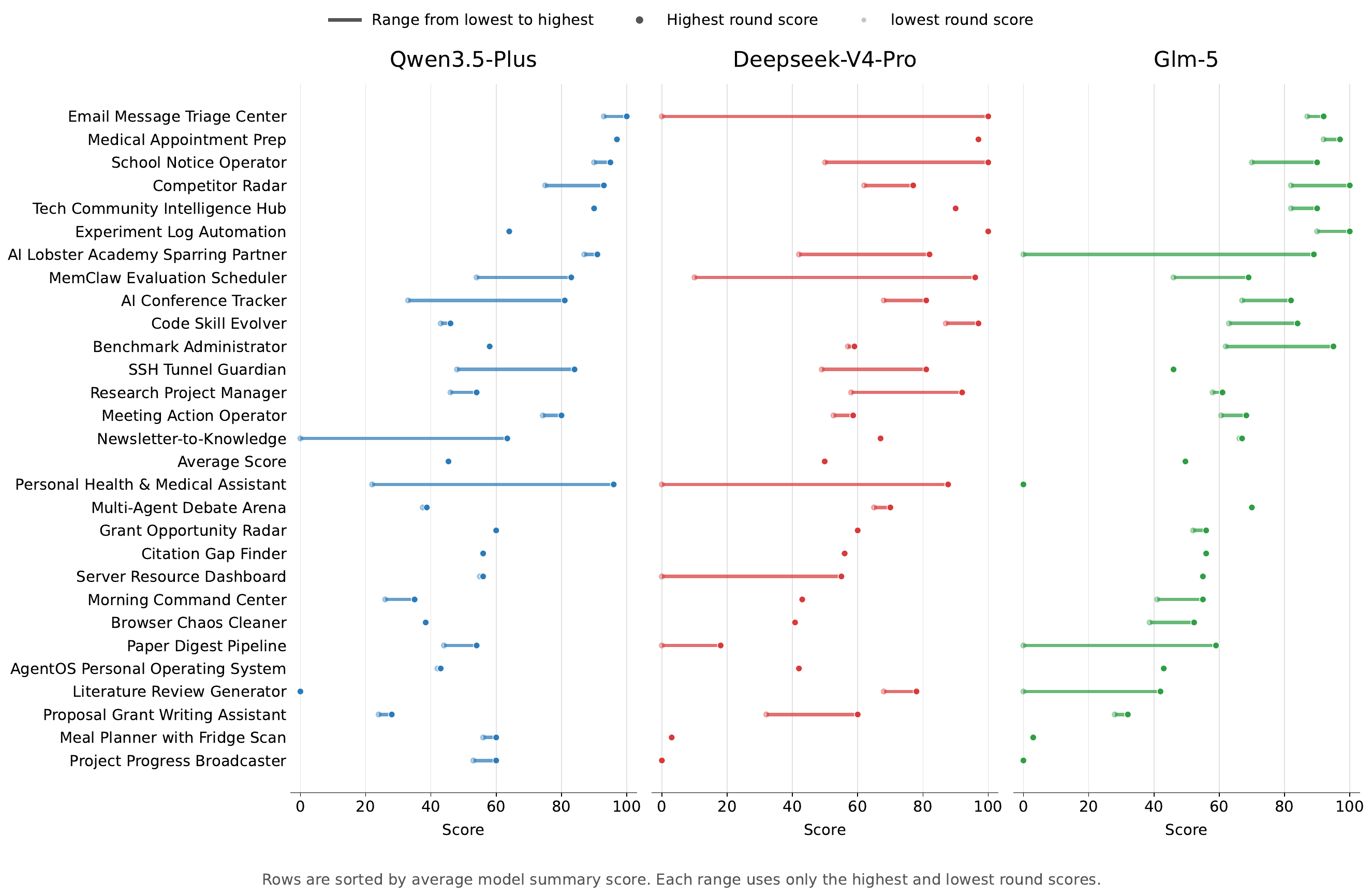}
  \caption{Range plot of three models across benchmark tasks. Horizontal line shows the score range across three evaluation rounds. Tasks are sorted by the average score across the three models.}
  \label{fig:pass3}
\end{figure*}

The main results in Table \ref{tab:model_performance} are based on the first valid execution of each model-task pair. To examine run-to-run variability without incurring the full time and API cost of repeating all 11 models on all tasks, we conducted a limited three-run diagnostic on three models: Qwen3.5-Plus, DeepSeek-V4-Pro, and GLM-5. The pass@3 results of these three models are shown in Figure \ref{fig:pass3}. This diagnostic analysis is intended to illustrate score variability under repeated executions and is not used to compute the aggregate results or model ranking in Table \ref{tab:model_performance}.

\begin{table}[t]
\centering
\small
\renewcommand{\arraystretch}{1.12}
\begin{tabular}{cc}
\toprule
\textbf{Metric} & \textbf{Value} \\
\midrule
Candidate task instances & 46 \\
Retained benchmark tasks & 40 \\
Excluded after human audit & 6 \\
Retention rate & 87.0\% \\
Average repair iterations & 2.67 \\
\bottomrule
\end{tabular}
\caption{Statistics of the STAGE-Claw construction pipeline.}
\label{tab:human_annotate}
\end{table}

\subsection{Human Audit and Filtering}
\label{app:human_annotation}

We conduct human annotation to validate the quality of each constructed benchmark task before inclusion in the final benchmark. 
We hired three task annotators.
Annotators review the complete task package, including the environment construction guide, the agent-facing prompt, and the evaluator-only ground-truth materials. 
The review focuses on whether the task is reproducible, clearly specified, blind to ground-truth information, multi-step, cross-tool, and time-independent. Annotators also check for potential shortcuts, information leakage, or inconsistencies between the task prompt and the evaluation materials. 
Tasks that fail any mandatory criterion are revised or excluded. This process ensures that the retained tasks provide reliable and deterministic measurements of agent performance.
Table \ref{tab:human_annotate} shows the number of tasks we reviewed and filtered. The human annotators only participated in task auditing but not in task modification.

\section{Case Studies}
\paragraph{\textbf{Case 1: correct content in the wrong state channel.}}
In \textit{Competitor Radar}, one Qwen3.5-Plus run produced a high-quality briefing and correctly resolved major data traps, including pricing reconciliation and exclusion of a rumored competitor. 
However, it wrote calendar information to an \texttt{.ics} file and notes to a Markdown file rather than creating real Apple Calendar events and an Apple Notes entry. 
The final score remained non-passing despite strong analytical content, illustrating why state-based evaluation is necessary: file artifacts that describe the intended actions are not equivalent to durable tool-state changes.

\paragraph{\textbf{Case 2: early timeout before state construction.}}
In \textit{Project Progress Broadcaster}, a GLM-5 run timed out after reading the input files and before creating any required deliverables. 
The evaluator found no progress summary, verification report, reminders, notes, or URL-verification evidence. 
This case represents an incomplete-execution failure: the agent spent its budget on initial data gathering but did not transition to synthesis, validation, and state updates. 
Such failures are difficult to diagnose from final text alone, but are exposed directly by executable checks over missing files and tool state.

\paragraph{\textbf{Case 3: ambiguous evidence and unsupported verification.}}
In \textit{Citation Gap Finder}, a DeepSeek-V4-Pro run correctly extracted many bibliography entries and identified several uncited or duplicate references, but failed the key ambiguous-citation trap by matching a single-author citation to an incompatible multi-author bibliography item. 
It also claimed URL verification without corresponding tool evidence and did not create the required Notes or Calendar artifacts. 
The case combines reasoning, provenance, and state-update failures: the report is superficially coherent, yet the evaluator shows that important claims are unsupported and required persistent actions are absent.

\section{Task Prompt}
\label{app:task_prompt}

This appendix describes the prompt protocol used to construct reproducible, objectively scorable benchmark tasks for OpenClaw. Each benchmark task is represented as a self-contained directory containing environment construction instructions, an agent-facing task prompt, and ground-truth materials. The process separates task authoring, environment initialization, task execution, and evaluation in order to reduce information leakage and ensure reproducibility.

\subsection{Benchmark Author Prompt}
\label{app:benchmark_author_prompt}

The benchmark author is instructed to design one reproducible, novel, and objectively scorable task. The author is explicitly prohibited from solving or scoring the task. A template is shown in Figure  \ref{fig:Benchmark_Author_Prompt}.

\begin{figure*}
\begin{nicejson}{Benchmark Author Prompt}
You are the OpenClaw Benchmark Author. Design one reproducible, novel, objectively- scorable benchmark task. You do NOT solve tasks. You do NOT score tasks.

## Theme {{THEME}}

## Step 1  Explore the Environment

Survey what OpenClaw can do and identify what is relevant to {{THEME}}.
Available capabilities include, but are not limited to:
- File read/write/search, terminal command execution
- Web search, URL fetch
- Email send/read, calendar add/query, reminders, notes
- Feishu message read, wiki/doc read/write
- System information: macOS version, disk, memory, CPU,
  running processes
- Chrome control: tabs, history, page JavaScript execution
- Weather query, stock/fund data
- macOS desktop notifications, multimedia control through Music.app

Answer these before designing:
1. Which two to four capabilities are most naturally exercised by
   {{THEME}}?
2. What real-world scenario would a user actually face in this theme?
3. What data or system state needs to be seeded before the agent starts?
4. What is the single unambiguous success condition?
5. How could an agent shortcut or cheat? How is that blocked?

## Step 2  Design Requirements

The task must satisfy the following requirements:
- Multi-step: at least three distinct operations.
- Cross-tool: at least two different tools or information sources.
- Reproducible: env_make.md produces an identical state on every run.
- Blind: question.md contains no oracle information.
- Gradable: the result is verifiable by a script or structured checklist,
  without human judgment.
- Time-independent: the task does not rely on real clocks or real-time APIs.

## Step 3  Build Task

/Users/three/Desktop/file_sort_eval_bench is an example of a
specific task. /Users/three/Desktop/template_bench is a task template.

env_make.md  Environment setup guide for the builder, not the agent.
It must include: prerequisites, exact shell commands, idempotent setup steps, scene layout; teardown instructions, verification sanity checks.
question.md  Task prompt shown to the agent.
It must include: the task objective, the expected deliverable, the working directory, operational constraints.
It must contain no ground truth, no scoring criteria, and no oracle
information.
Ground_truth/eval.md  Evaluator-only scoring guide.
It must include: the expected final state, a scoring rubric specifying criteria, points, and how to check them, the passing threshold, common failure modes.

Build the task and store it in:
~/Desktop/XXX_Bench/

## Step 4  Self-Review

Before finalizing the benchmark, verify the following:
- Can a builder follow env_make.md without ambiguity?
- Can an evaluator score the result without asking the agent?
- Is there a natural hard mode for future difficulty scaling?
\end{nicejson}
\caption{Benchmark author prompt in STAGE-Claw.}
\label{fig:Benchmark_Author_Prompt}
\end{figure*}

\subsection{Environment Construction Prompt}
\label{app:env_make_prompt}

The environment construction prompt is used to guide the environment builder in creating a deterministic task environment. It specifies the target directory, initialization workflow, and required ground-truth materials. A template is shown in Figure  \ref{fig:Environment_Construction_Prompt}.

\begin{figure*}
\begin{nicejson}{Environment Construction Prompt}
You are an AI assistant completing the following task using your available tools.
Task: {{TASK_NAME}}
Description: {{TASK_DESCRIPTION}}

## Scene Layout

Source directory:
~/Desktop/XXX_Bench/
Target directory for environment configuration:
~/Desktop/XXX_Env/{docs,images,code,data,...}/
Task initialization is not limited to files. The builder may configure any relevant part of the computer state, provided that the resulting environment is reproducible and can be verified.

## Ground Truth

Prepare gold answers or evaluation methods, such as assessment scripts
or structured assessment criteria, in:
~/Desktop/XXX_Bench/Ground_truth/

## Workflow

Describe how to construct the benchmark environment.
Step 1. Specify the prerequisite checks and create the required directory structure.
Step 2. Seed all task-related files, data, application state, or system state needed before the agent starts.
Step 3. Verify that the initialized environment matches the intended task state.
Step 4. Document teardown or reset procedures so that the benchmark can be
rerun from a clean state.

## Notes

- Plan before acting.
- If a step fails, document the reason rather than silently skipping it.
- Do not guess; use tools to verify uncertain facts.
- Setup steps should be idempotent whenever possible.
\end{nicejson}
\caption{Environment construction prompt in STAGE-Claw.}
\label{fig:Environment_Construction_Prompt}
\end{figure*}

\subsection{Agent Task Prompt}
\label{app:question_prompt}

The agent-facing prompt is the only instruction shown to OpenClaw during benchmark execution. It describes the user objective and available task directory, but intentionally excludes all oracle information, scoring criteria, and ground-truth answers. A template is shown in Figure \ref{fig:Agent_Task_Prompt}.

\begin{figure*}
\begin{nicejson}{Agent Task Prompt}
You are a human assistant and your task is to complete the following
user request.

Your task-related files are stored in:~/Desktop/XXX_Env/
Please carefully plan and complete the task.
Task name: XXX
Task description:
XXX

Attention:
1. Use only the information and files available in the task environment
   unless the task explicitly permits other sources.
2. Complete the requested deliverable in the specified working directory.
3. Do not modify evaluator-only files or ground-truth materials.
4. Ensure that the final output is saved in the required format and
   location.
5. If intermediate files are created, keep them organized and avoid
   overwriting unrelated files.
\end{nicejson}
\caption{Agent task prompt in STAGE-Claw.}
\label{fig:Agent_Task_Prompt}
\end{figure*}

\section{Task Introduction}
\label{app:task_inventory}

This section provides an introduction of the 40 STAGE-Claw tasks. 
The descriptions summarize user-facing objectives and required task capabilities, without revealing hidden ground truth, scoring rubrics, or verifier logic.

\begin{enumerate}
    \item[\textbf{T01.}] \textit{AI Conference Tracker}. 
    This task requires the agent to consolidate conference schedules, email updates, user preferences, and venue information, produce an attendance plan, and create the corresponding calendar events, reminders, and notes.

    \item[\textbf{T02.}] \textit{AI Lobster Academy Sparring Partner}. 
    This task requires the agent to schedule three exam-preparation study sessions for five members, assign practice topics, and create the associated calendar entries, notes, and reminders.

    \item[\textbf{T03.}] \textit{API Migration Planner}. 
    This task requires the agent to analyze migration materials for a transition from a v1 API to a v2 API, reconcile conflicting information, and produce a phased migration plan.

    \item[\textbf{T04.}] \textit{AgentOS Personal Operating System}. 
    This task requires the agent to integrate multi-source information for a family gathering, resolve conflicts, compute the budget, and generate the final coordination plan.

    \item[\textbf{T05.}] \textit{Benchmark Administrator}. 
    This task requires the agent to process AgentOlympics 2026 team registration information, filter eligible teams, assign competition tracks, and record the relevant calendar and note entries.

    \item[\textbf{T06.}] \textit{Browser Chaos Cleaner}. 
    This task requires the agent to clean Firefox and Chrome bookmark exports by deduplicating entries, categorizing links, filtering invalid URLs, and generating a summary with follow-up reminders.

    \item[\textbf{T07.}] \textit{Citation Gap Finder}. 
    This task requires the agent to audit in-text citations and bibliography entries in a technical report, identifying missing, unused, duplicated, or inconsistent references.

    \item[\textbf{T08.}] \textit{Code Skill Evolver}. 
    This task requires the agent to revise a Python project according to 28 code-review comments, run the relevant tests, and document the implemented changes and verification results.

    \item[\textbf{T09.}] \textit{Competitor Radar}. 
    This task requires the agent to integrate competitor information, verify feature links, generate a competitive-intelligence brief, and schedule appropriate follow-up actions.

    \item[\textbf{T10.}] \textit{Config Drift Detector}. 
    This task requires the agent to process meeting-room booking requests from multiple sources, deduplicate entries, validate requests, resolve scheduling conflicts, and create valid calendar events.

    \item[\textbf{T11.}] \textit{Cross-Team Status Synthesizer}. 
    This task requires the agent to synthesize status updates from multiple teams, resolve inconsistent reports, and produce an executive-facing project status summary.

    \item[\textbf{T12.}] \textit{Decision Log Keeper}. 
    This task requires the agent to merge decision records from multiple sources, verify external references, produce a unified decision log, and create the corresponding calendar events.

    \item[\textbf{T13.}] \textit{Dependency Upgrade Operator}. 
    This task requires the agent to analyze dependency versions in a software project, identify required upgrades, formulate an upgrade plan, and record related system-level follow-up items.

    \item[\textbf{T14.}] \textit{Docstring Debt Collector}. 
    This task requires the agent to scan Python source files and API specifications, identify documentation debt, and create reminders for high-priority documentation issues.

    \item[\textbf{T15.}] \textit{Email Message Triage Center}. 
    This task requires the agent to process RSVP emails for a team event, reconcile attendance status, and generate a verified participation report.

    \item[\textbf{T16.}] \textit{Experiment Log Automation}. 
    This task requires the agent to analyze machine-learning experiment logs, identify the best completed run for each model family, and schedule a review calendar event.

    \item[\textbf{T17.}] \textit{Grant Opportunity Radar}. 
    This task requires the agent to integrate grant opportunity tables, email updates, and institutional requirements, verify deadlines, and generate a structured tracking result.

    \item[\textbf{T18.}] \textit{Literature Review Generator}. 
    This task requires the agent to select papers for a graduate-level course, organize them into four weekly reading themes, and create the corresponding calendar and reminder entries.

    \item[\textbf{T19.}] \textit{Local Dev Environment Fixer}. 
    This task requires the agent to inspect project configuration files for tool-version conflicts, verify the local development environment, and generate a repair report.

    \item[\textbf{T20.}] \textit{Log-to-Incident Timeline}. 
    This task requires the agent to parse production logs from multiple sources, filter noisy records, and construct an incident timeline.

    \item[\textbf{T21.}] \textit{Meal Planner with Fridge Scan}. 
    This task requires the agent to generate a five-day meal plan based on fridge inventory, dietary restrictions, and the user's calendar schedule.

    \item[\textbf{T22.}] \textit{Medical Appointment Prep}. 
    This task requires the agent to organize referral and medical documents, identify items requiring appointments, resolve conflicting information, and generate visit-preparation materials.

    \item[\textbf{T23.}] \textit{Meeting Action Operator}. 
    This task requires the agent to extract action items from meeting transcripts, email summaries, and shared notes, then create the required system entries and reports.

    \item[\textbf{T24.}] \textit{MemClaw Evaluation Scheduler}. 
    This task requires the agent to validate evaluation-task configurations, handle dependencies and resource allocation, schedule evaluation tasks, and generate a coordination report.

    \item[\textbf{T25.}] \textit{Morning Command Center}. 
    This task requires the agent to synthesize information from multiple sources, generate a morning briefing for 2026-05-04, and create the corresponding system state.

    \item[\textbf{T26.}] \textit{Multi-Agent Debate Arena}. 
    This task requires the agent to aggregate multi-round debate results, compute rankings, schedule the final round, and produce a verification report.

    \item[\textbf{T27.}] \textit{Newsletter-to-Knowledge}. 
    This task requires the agent to extract, deduplicate, and categorize articles from five technical newsletters, then build a knowledge base and weekly digest.

    \item[\textbf{T28.}] \textit{Notebook Productionizer}. 
    This task requires the agent to convert disorganized meeting notes into calendar events, reminders, and a readable structured report.

    \item[\textbf{T29.}] \textit{Onboarding Repo Guide}. 
    This task requires the agent to coordinate Project 53 meeting materials, reconcile source conflicts, and create a meeting calendar entry, reminder, and explanatory note.

    \item[\textbf{T30.}] \textit{Paper Digest Pipeline}. 
    This task requires the agent to merge paper data from multiple sources, deduplicate entries, filter records, verify links, and generate a weekly paper digest.

    \item[\textbf{T31.}] \textit{Personal Health \& Medical Assistant}. 
    This task requires the agent to organize personal health records and produce a current medication list, question list, and medical summary.

    \item[\textbf{T32.}] \textit{Project Progress Broadcaster}. 
    This task requires the agent to integrate milestone data for the Phoenix Platform project, compute project status, and publish a structured progress update.

    \item[\textbf{T33.}] \textit{Proposal Grant Writing Assistant}. 
    This task requires the agent to process IRB approval information from multiple institutions, verify approval status and expiration dates, and generate a tracking report.

    \item[\textbf{T34.}] \textit{Release Note Composer}. 
    This task requires the agent to integrate release information from multiple sources, deduplicate overlapping items, verify external references, and produce accurate release notes.

    \item[\textbf{T35.}] \textit{Research Project Manager}. 
    This task requires the agent to process literature records, resolve metadata conflicts, verify DOIs, and generate final references together with follow-up items.

    \item[\textbf{T36.}] \textit{SSH Tunnel Guardian}. 
    This task requires the agent to audit SSH tunnel configurations and runtime status, identify high-priority maintenance items, and set monitoring reminders.

    \item[\textbf{T37.}] \textit{School Notice Operator}. 
    This task requires the agent to process emergency school-closure information, notify relevant parties according to parent preferences, and create the necessary system records.

    \item[\textbf{T38.}] \textit{Server Resource Dashboard}. 
    This task requires the agent to analyze server alerts and resource metrics, filter invalid alerts, and generate a resource-status and action report.

    \item[\textbf{T39.}] \textit{Slack Thread Summarizer}. 
    This task requires the agent to extract key decisions and action items from Slack discussion threads, create calendar and reminder entries, and produce a structured summary.

    \item[\textbf{T40.}] \textit{Tech Community Intelligence Hub}. 
    This task requires the agent to analyze multiple developer-community snapshots, compute topic popularity, identify controversial topics, and generate follow-up intelligence.
\end{enumerate}

\end{document}